%% file: ms.tex
\documentclass[final]{cvpr}

\usepackage{times}
\usepackage{epsfig}
\usepackage{graphicx}
\usepackage{amsmath}
\usepackage{amssymb}
\usepackage{multirow}
\usepackage{caption,subcaption}
\usepackage{floatrow}

\usepackage[dvipsnames]{xcolor}

\usepackage{color}
\usepackage{soul}

\usepackage[pagebackref=true,breaklinks=true,colorlinks,bookmarks=false]{hyperref}

\def\cvprPaperID{2631} 
\def\confYear{CVPR 2021}
\definecolor{myPurple}{rgb}{0.4, .0, .8}
\definecolor{myGreen}{rgb}{0, .8, .3}
\definecolor{myRed}{rgb}{0.8, .2, .2}
\definecolor{myOrange}{rgb}{0.8, 0.45, 0.0}
\definecolor{myBlue}{rgb}{.0, .0, 1.0}
\definecolor{myBlue2}{rgb}{.0, .0, 0.5}
\definecolor{myBlack}{rgb}{.0, .0, 0.0}
\definecolor{darkmidnightblue}{rgb}{0.0, 0.2, 0.4}

\begin{document}

\title{Learning Camera Localization via Dense Scene Matching}
\author{Shitao Tang$^{1}$ \quad {Chengzhou Tang}$^{1}$ \quad {Rui Huang}$^{2}$ \quad {Siyu Zhu}$^{2}$ \quad {Ping Tan}$^{1}$ \\
	${^1}$Simon Fraser University \quad ${^2}$Alibaba A.I Labs \\
	{\tt\small \{shitao\_tang, chengzhou\_tang, pingtan\}@sfu.ca} \quad {\tt\small \{rui.hr, siting.zsy\}@alibaba-inc.com}
	
}


\maketitle
\begin{abstract}
Camera localization aims to estimate 6 DoF camera poses from RGB images. Traditional methods detect and match interest points between a query image and a pre-built 3D model. Recent learning-based approaches encode scene structures into a specific convolutional neural network (CNN) and thus are able to predict dense coordinates from RGB images. However, most of them require re-training or re-adaption for a new scene and have difficulties in handling large-scale scenes due to limited network capacity. We present a new method for scene agnostic camera localization using dense scene matching (DSM), where a cost volume is constructed between a query image and a scene. The cost volume and the corresponding coordinates are processed by a CNN to predict dense coordinates. Camera poses can then be solved by PnP algorithms. In addition, our method can be extended to temporal domain, which leads to extra performance boost during testing time. Our scene-agnostic approach achieves comparable accuracy as the existing scene-specific approaches, such as KFNet, on the 7scenes and Cambridge benchmark. This approach also remarkably outperforms state-of-the-art scene-agnostic dense coordinate regression network SANet. The Code is available at \href{https://github.com/Tangshitao/Dense-Scene-Matching}{https://github.com/Tangshitao/Dense-Scene-Matching}.

\end{abstract}

\input{introduction}

\input{relatedwork}
\input{3_1_overview}
\input{3_2_pyramid}
\input{3_3_dense_matching}

\input{3_4_loss}

\input{table1}
\input{table2}
\input{experiment}
\input{conclusion}

\appendix

\section{Archtecture of $Net_{coords}$ and $Net_{conf}$.}
\begin{figure}
    \centering
    \includegraphics[width=\textwidth]{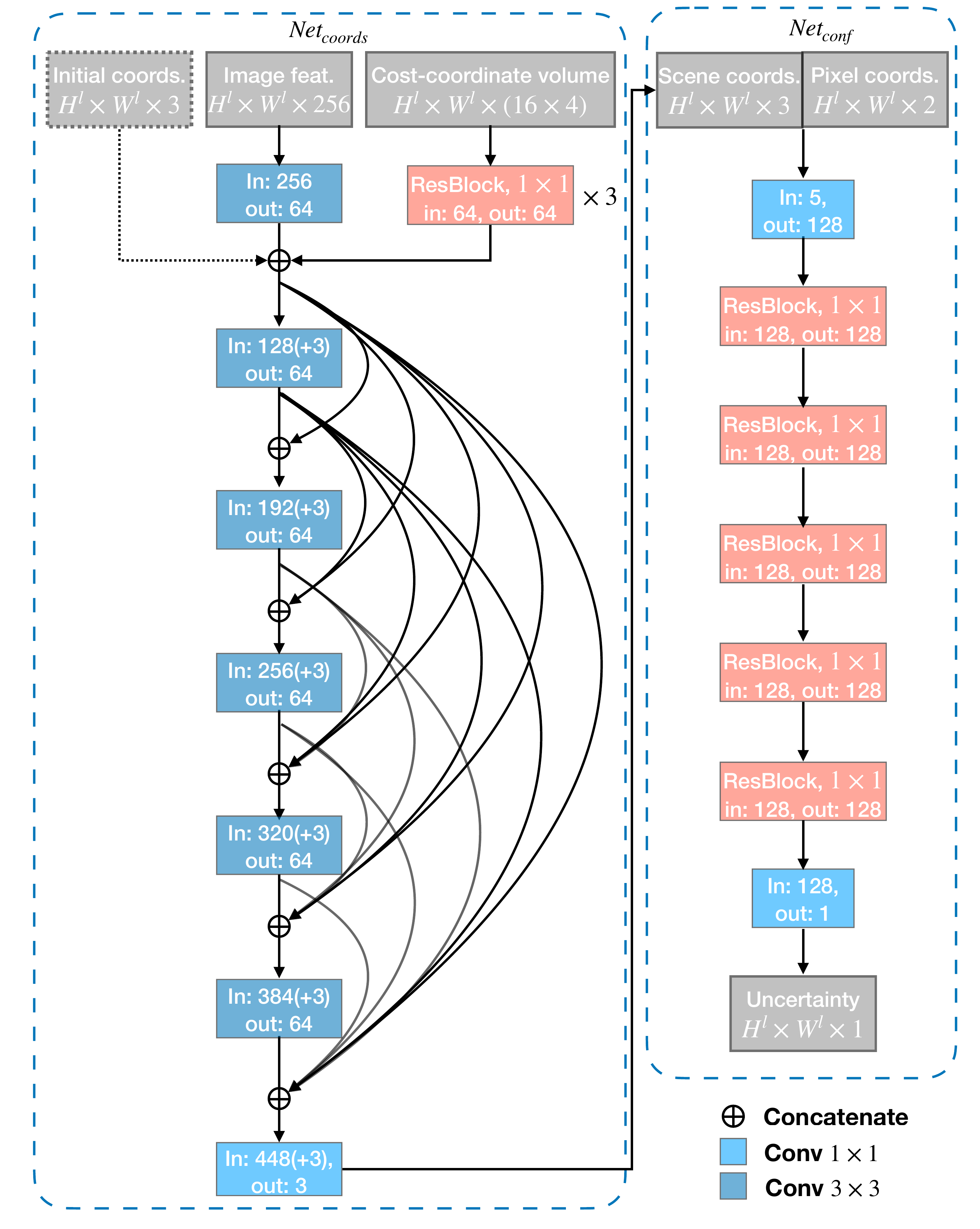}
    \caption{Archtecture of $Net_{conf}$ and $Net_{coords}$. We use residual block for $Net_{conf}$ and dense block for $Net_{coords}$}
    \label{fig:arch}
\end{figure}

Fig. \ref{fig:arch} shows the architecture of $Net_{conf}$ and $Net_{coords}$. The input of $Net_{coods}$ is a $H^l \times W^l \times 4K$ ($K=16$ in implementation) cost-coordinate volume formed by concatenating the cost volume with 3D scene coordinates. As shown in Fig. \ref{fig:arch}, $Net_{coods}$ consists of 3 residual blocks~\cite{he2016deep} and one denseblock~\cite{huang2017densely}. The residual blocks consist of $1\times 1$ convolutional layer. It takes input of cost-coordinate volume and generates a $H^l \times W^l \times 64$ coordinate feature map. Then, the scene coordinate map is estimated by the denseblock, which takes the concatenation of image features, coordinate features and the initial coordinate map up-sampled from last layer (if applicable). On the other hand, $Net_{conf}$ consists of 5 residual block with context normalization~\cite{yi2018learning}. It takes the concatenation of the estimated scene coordinate map with the corresponding 2D pixel coordinate map and estimates a confidence score for each pixel.

\section{Additional Analysis}
This section provides additional analysis of DSM. All the experiments are conducted on 7scenes dataset. The data processing and training process are the same as described in the main paper. At the inference time, We use 1 out of every 10 frames for each sequence. Pose accuracy, the percentage of predicted poses falling within the threshold (5$^{\circ}$, 5cm), is used as the evaluation metric.

\noindent \textbf{Effects of correlation sorting.} As described in Sec.3.3.1 of the main paper, one of the procedures in cost volume construction is sorting and selecting top $K$ coordinates for each pixel from the correlation tensor. The motivation behind this operation is two-fold. Firstly, as the number of retrieved scene images varies, top $K$ selection results in a cost volume with a fixed size. Secondly, a sorted cost volume leads to a more accurate estimated coordinate map. To verify the effectiveness of correlation sorting, we fix the scene image number to 5 and directly use the correlation tensor as the cost volume for coordinate map regression. The results are shown in Table. \ref{tab:sort}. It can be seen that the estimated pose accuracy improves by correlation sorting consistently on all sequences. Moreover, since top $K$ sorting and selection results in a fixed-size cost volume, we can use different scene image numbers for training and testing. During the training process, the scene image number can be fixed for better efficiency while for inference we can leverage more scene images for higher accuracy.

\begin{table}
    \centering
    \scalebox{0.75}{\begin{tabular}{c|c|c|c|c|c|c|c}
    \hline
    &Chess&Fire&Heads&Office&Pumpkin&Kitchen&Stairs\\
    \hline
      No sorting&0.82&0.74&0.85&0.72&0.43&0.58&0.05  \\

        Sorting&0.96&0.95&1.0&0.88&0.53&0.72&0.66 \\
    \hline
    \end{tabular}
    \caption{Pose accuracy with/without top $K$ correlation sorting. The estimated pose accuracy improves by correlation sorting consistently on all sequences.}
    \label{tab:sort}}
\end{table}

\begin{table}[]
    \centering
    \scalebox{0.75}{\begin{tabular}{c|c|c|c|c|c|c|c}
    \hline
    Num.&Chess&Fire&Heads&Office&Pumpkin&Kitchen&Stairs\\
    \hline
       1&0.87&0.85&0.87&0.71&0.45&0.63&0.17  \\
        3&0.90&0.94&0.91&0.79&0.46&0.67&0.20 \\
        5&0.94&0.94&0.94&0.80&0.54&0.68&0.24\\
        10 ($\star$)&0.96&0.95&1.0&0.88&0.53&0.72&0.66 \\
    \hline
    \end{tabular}
    \caption{Pose accuracy with respect to the number of scene images. The network is trained and tested with the corresponding number of scene images except the one with 10 scene images. The notation ($\star$) means we train the network with 5 scene images instead of 10 scene images.}
    \label{tab:num}}
\end{table}

\begin{table}
    \centering
    \scalebox{0.75}{\begin{tabular}{c|c|c|c|c|c|c|c}
    \hline
    Reso.&Chess&Fire&Heads&Office&Pumpkin&Kitchen&Stairs\\
    \hline
      $192\times 256$&0.92&0.84&0.89&0.78&0.49&0.64&0.23   \\

        $384\times 512$&0.96&0.95&1.0&0.88&0.53&0.72&0.66 \\
    \hline
    \end{tabular}
    \caption{Pose accuracy with respect to different image resolutions. In our implementation, We resize all images to resolution of $384\times 512$ for better efficiency and performance.}
    \label{tab:reso}}
\end{table}

\noindent \textbf{Number of scene image.} To show the effects of scene image number $N$, we change $N$ from 1 to 10 in the training and testing process to evaluate the pose accuracy. The model is re-trained with respect to the corresponding scene image number for $N=1,3,5$. Since training with more than 5 scene images leads to unacceptable GPU memory consumption, we still use 5 scene images in training when testing with 10 scene images. As shown in Table~\ref{tab:num}, increasing $N$ from $1$ to $5$ results in higher pose accuracy. In addition, we can see that 10 scene images obtain higher performance than 5 scene images. This indicates that even if the model is trained with fewer scene images, leveraging more scene images leads to better performance. Considering the trade-off between performance and efficiency, we set $N=10$ in the main paper.

\noindent \textbf{Image resolution.} We test our model using 2 different image resolution size $192\times 256$ and $384\times 512$. As shown in Table. \ref{tab:reso}, we can see the resolution of $384\times 512$ outperforms $192\times 256$. A higher resolution than $384\times 512$ could consume more GPU memory and lead to slower running time. Therefore, we resize all the images to $384\times 512$ in our system for better efficiency.

\begin{figure*}
    \centering 
\begin{subfigure}{0.18\textwidth}
  \includegraphics[width=\linewidth]{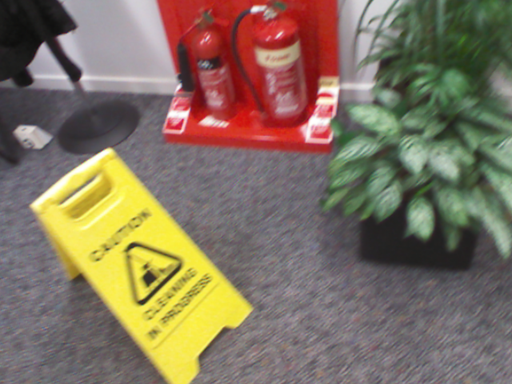}
\end{subfigure}\hfil 
\begin{subfigure}{0.18\textwidth}
  \includegraphics[width=\linewidth]{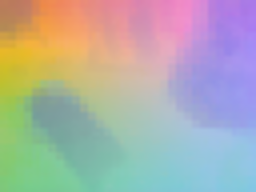}
\end{subfigure}\hfil 
\begin{subfigure}{0.18\textwidth}
  \includegraphics[width=\linewidth]{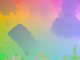}
\end{subfigure}\hfil 
\begin{subfigure}{0.18\textwidth}
  \includegraphics[width=\linewidth]{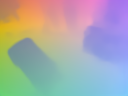}
\end{subfigure}\hfil 
\begin{subfigure}{0.18\textwidth}
  \includegraphics[width=\linewidth]{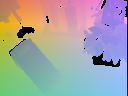}
\end{subfigure}\hfil 

\begin{subfigure}{0.18\textwidth}
  \includegraphics[width=\linewidth]{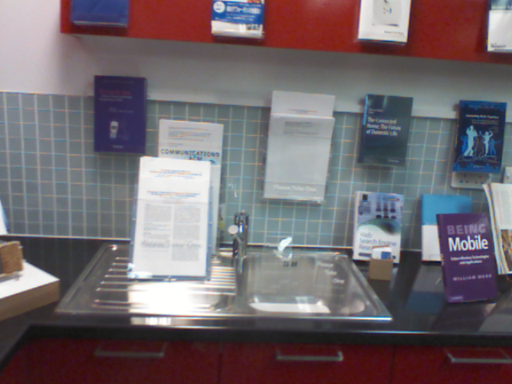}
\end{subfigure}\hfil 
\begin{subfigure}{0.18\textwidth}
  \includegraphics[width=\linewidth]{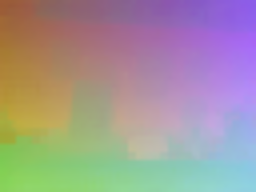}
\end{subfigure}\hfil 
\begin{subfigure}{0.18\textwidth}
  \includegraphics[width=\linewidth]{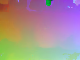}
\end{subfigure}\hfil 
\begin{subfigure}{0.18\textwidth}
  \includegraphics[width=\linewidth]{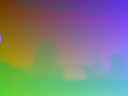}
\end{subfigure}\hfil 
\begin{subfigure}{0.18\textwidth}
  \includegraphics[width=\linewidth]{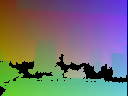}
\end{subfigure}\hfil 

\begin{subfigure}{0.18\textwidth}
  \includegraphics[width=\linewidth]{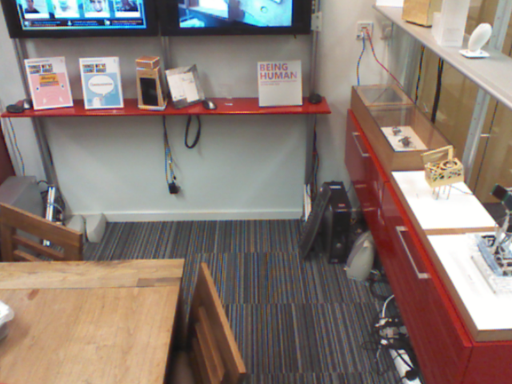}
\end{subfigure}\hfil 
\begin{subfigure}{0.18\textwidth}
  \includegraphics[width=\linewidth]{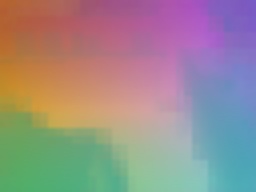}
\end{subfigure}\hfil 
\begin{subfigure}{0.18\textwidth}
  \includegraphics[width=\linewidth]{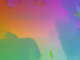}
\end{subfigure}\hfil 
\begin{subfigure}{0.18\textwidth}
  \includegraphics[width=\linewidth]{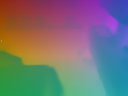}
\end{subfigure}\hfil 
\begin{subfigure}{0.18\textwidth}
  \includegraphics[width=\linewidth]{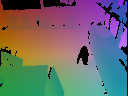}
\end{subfigure}\hfil 

\begin{subfigure}{0.18\textwidth}
  \includegraphics[width=\linewidth]{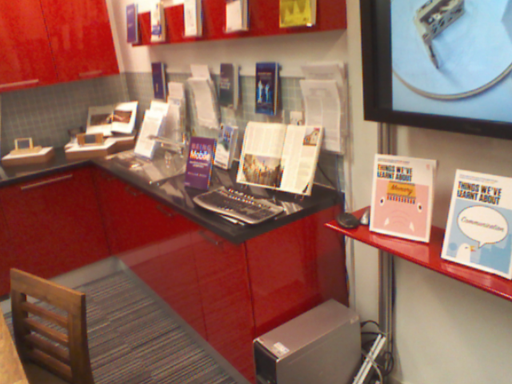}
\end{subfigure}\hfil 
\begin{subfigure}{0.18\textwidth}
  \includegraphics[width=\linewidth]{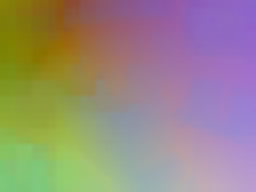}
\end{subfigure}\hfil 
\begin{subfigure}{0.18\textwidth}
  \includegraphics[width=\linewidth]{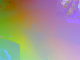}
\end{subfigure}\hfil 
\begin{subfigure}{0.18\textwidth}
  \includegraphics[width=\linewidth]{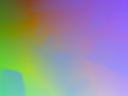}
\end{subfigure}\hfil 
\begin{subfigure}{0.18\textwidth}
  \includegraphics[width=\linewidth]{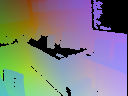}
\end{subfigure}\hfil 

\begin{subfigure}{0.18\textwidth}
  \includegraphics[width=\linewidth]{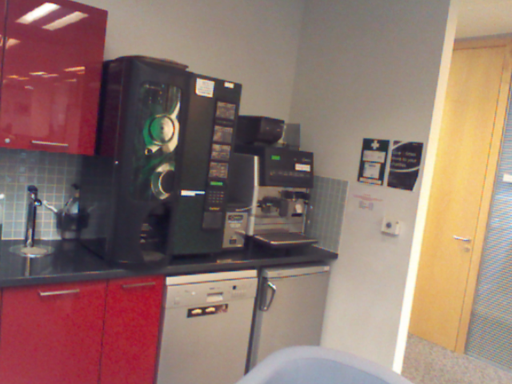}
\end{subfigure}\hfil 
\begin{subfigure}{0.18\textwidth}
  \includegraphics[width=\linewidth]{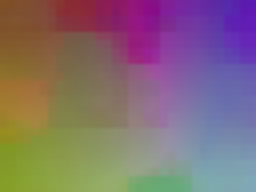}
\end{subfigure}\hfil 
\begin{subfigure}{0.18\textwidth}
  \includegraphics[width=\linewidth]{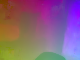}
\end{subfigure}\hfil 
\begin{subfigure}{0.18\textwidth}
  \includegraphics[width=\linewidth]{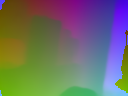}
\end{subfigure}\hfil 
\begin{subfigure}{0.18\textwidth}
  \includegraphics[width=\linewidth]{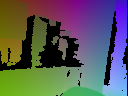}
\end{subfigure}\hfil 

\begin{subfigure}{0.18\textwidth}
  \includegraphics[width=\linewidth]{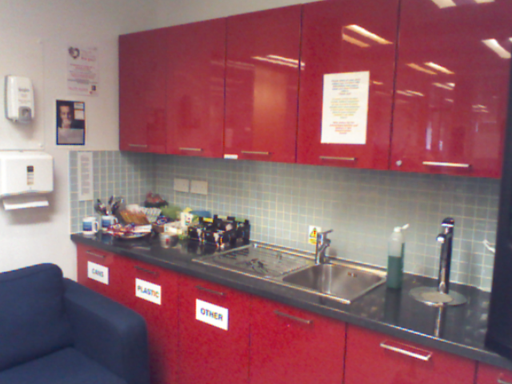}
\end{subfigure}\hfil 
\begin{subfigure}{0.18\textwidth}
  \includegraphics[width=\linewidth]{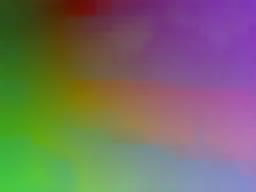}
\end{subfigure}\hfil 
\begin{subfigure}{0.18\textwidth}
  \includegraphics[width=\linewidth]{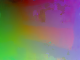}
\end{subfigure}\hfil 
\begin{subfigure}{0.18\textwidth}
  \includegraphics[width=\linewidth]{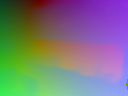}
\end{subfigure}\hfil 
\begin{subfigure}{0.18\textwidth}
  \includegraphics[width=\linewidth]{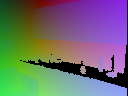}
\end{subfigure}\hfil 

\begin{subfigure}{0.18\textwidth}
  \includegraphics[width=\linewidth]{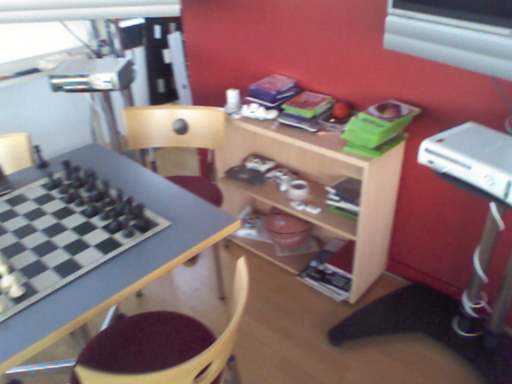}
\caption{Query}
\end{subfigure}\hfil 
\begin{subfigure}{0.18\textwidth}
  \includegraphics[width=\linewidth]{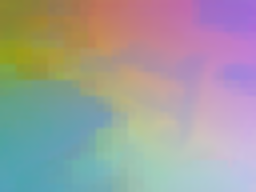}
  \caption{SANet}\label{fig:2}
\end{subfigure}\hfil 
\begin{subfigure}{0.18\textwidth}
  \includegraphics[width=\linewidth]{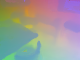}
  \caption{DSAC++}
\end{subfigure}\hfil 
\begin{subfigure}{0.18\textwidth}
  \includegraphics[width=\linewidth]{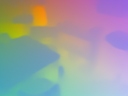}
   \caption{DSM}\label{DSM_A}
\end{subfigure}\hfil 
\begin{subfigure}{0.18\textwidth}
  \includegraphics[width=\linewidth]{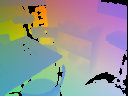}
\caption{G.T.}
\end{subfigure}\hfil 

\caption{Coordinate map visualization for SANet, DSAC++ and DSM. The coordinate maps produced by DSM recover more details as in the ground truth and have fewer artifacts than SANet and DSAC++.}\label{vis}
\end{figure*}

\noindent \textbf{More coordinate map visualization.} Fig. \ref{vis} provides more visualization results on the comparison of estimated coordinate maps from SANet, DASC++, and our method (DSM). In general, the coordinate maps produced by DSM recover more details as in the ground truth and have fewer artifacts than SANet and DSAC++.


{\small
\bibliographystyle{ieee_fullname}
\bibliography{egbib}
}

\end{document}

%% file: introduction.tex
\section{Introduction}
Camera Localization aims to estimate a 6-DoF camera pose of an image in a known environment. It is an important module in applications such as mobile navigation, simultaneous localization and mapping (SLAM) and augmented reality (AR). Camera localization methods can be broadly categorized as regression-based and structure-based. Earlier methods~\cite{kendall2015posenet,kendall2016modelling,kendall2017geometric, walch2017image} directly regress the camera poses from images, which are limited by the nature of image retrieval and generally less accurate~\cite{sattler2019understanding}. In comparison, structure-based methods~\cite{brachmann2017dsac, schohn2000less, brachmann2019expert, sarlin2019coarse, zhou2020kfnet, sattler2012improving, taira2018inloc} gradually become the trend and solve the problem in two stages: first, establishing the correspondences between 2D query image pixels and 3D scene points; second, estimating the desired camera pose by PnP~\cite{hesch2011direct} combined with different RANSAC~\cite{fischler1981random} algorithms. 

According to how they establish the 2D-3D correspondences, the structure-based methods can be further categorized into two classes: 1) sparse feature matching~\cite{sarlin2019coarse, sarlin2020superglue,sattler2012improving,taira2018inloc}; 2) scene coordinate map regression~\cite{brachmann2017dsac, schohn2000less, brachmann2019expert, zhou2020kfnet,li2020hierarchical}. The sparse feature matching methods detect and match handcrafted~\cite{lowe2004distinctive} or CNN-based~\cite{detone2018superpoint, sarlin2020superglue} feature points between a query image and scene images, which is able to handle arbitrary scenes. On the other hand, coordinate map regression methods predict dense 3D coordinates at all image pixels from a random forest~\cite{shotton2013scene} or a convolutional neural network (CNN)~\cite{brachmann2017dsac,schohn2000less}. The estimated dense coordinate maps can be effectively applied to augmented reality and robotics applications such as virtual object insertion or obstacle avoidance. But these methods are often limited to the scene where the random forests or CNN is trained. 

In this paper, we focus on the coordinate map regression approach. Recently, instead of encoding specific scene information in network parameters~\cite{shotton2013scene,brachmann2017dsac,schohn2000less}, Yang et.al, propose the first dense coordinate regression network SANet for arbitrary scenes~\cite{yang2019sanet}. SANet extracts a scene representation from some scene images and corresponding 3D coordinates by 2D-3D matching. In this way, it can be applied to different scenes without re-training or re-adaption. However, due to the irregular nature of a scene, SANet randomly selects coordinates within a region using ball query and leverages PointNet~\cite{qi2017pointnet} to regress per-pixel 3D coordinates. This operation undermines the pose accuracy and is computationally-heavy because a shared PointNet is required to make prediction on each pixel individually. 

\begin{figure*}
    \centering
    \includegraphics[width=0.95\textwidth]{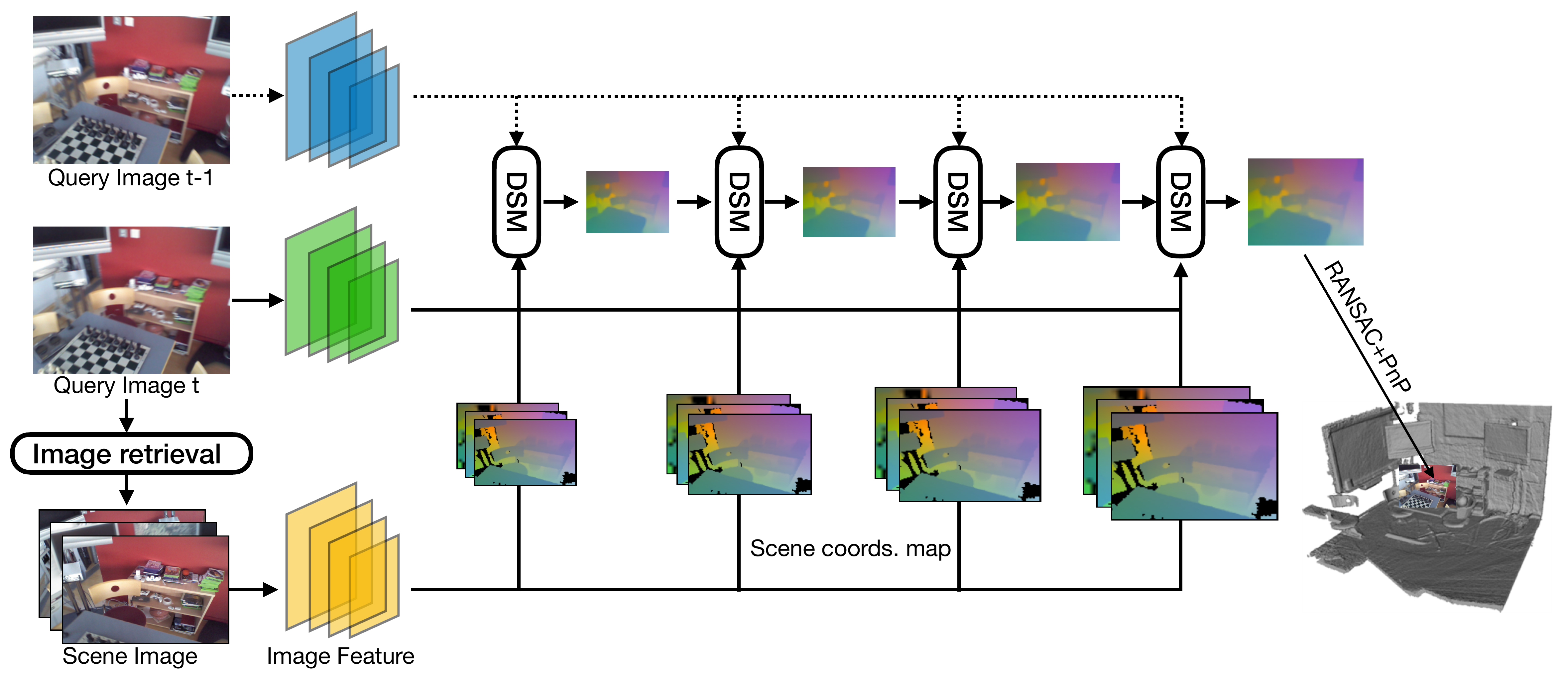}
    \caption{Overview of our framework. Our method predicts dense coordinate maps in a coarse-to-fine manner. The DSM module receives a query image feature map, some scene image feature maps and the corresponding scene coordinates to predict a dense coordinate map for the query image. This predicted scene coordinates are then used to solve camera poses with RANSAC and PnP algorithms.}
    \label{fig:pipeline}
\end{figure*}

In order to address this problem, we present a new scene-agnostic camera localization network exploiting dense scene matching (DSM), which matches each query image pixel with the scene via a cost volume. With end-to-end training, the cost volume explicitly enforces more accurate scene points to have a higher correlation with the input query pixel. Since the scene structure is irregular, which makes the number of query-scene correlations different for each image pixel, we propose a simple yet effective solution to unify the size of~\emph{all} cost volumes: sorting and selecting the best $K$ candidates and feed them to a convolutional neural network for dense coordinate regression. The cost volume can be further fused with temporal correlations between consecutive query images during inference, so that our method can be extended to video localization.  

We have evaluated our method on several benchmark datasets including indoor scenes, 7scenes~\cite{shotton2013scene} and large-scale outdoor scenes, Cambridge~\cite{kendall2015posenet}. We have shown DSM achieves state-of-the-art performance among scene-specific methods including DSAC++ in terms of both pose accuracy and coordinate accuracy, and outperforms scene-agnostic methods, e.g. SANet, by a large margin.

%% file: relatedwork.tex
\begin{figure*}
    \centering
    \includegraphics[width=0.95\textwidth]{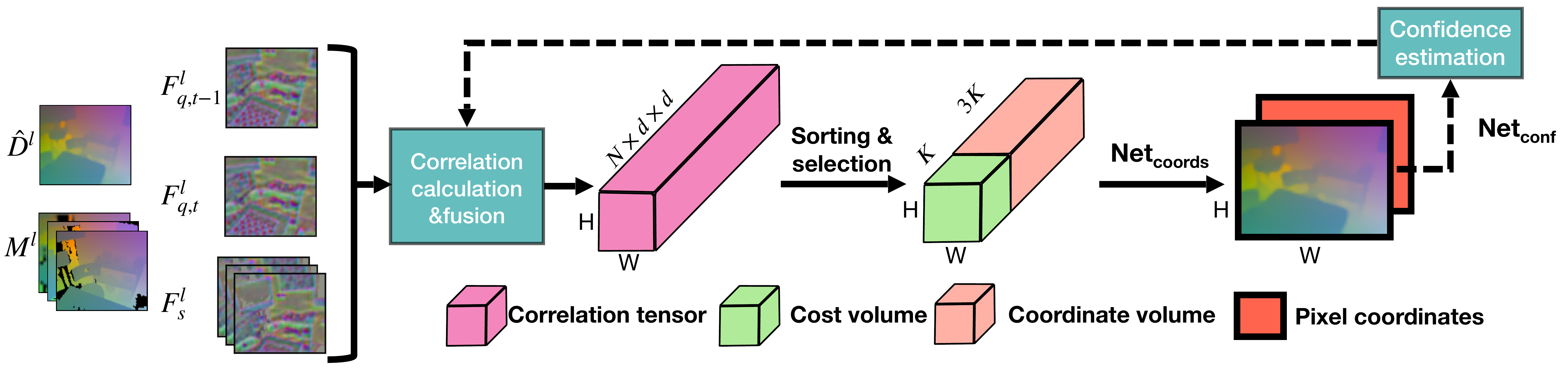}
    \caption{Illustration of Dense Scene Matching (DSM) module. For  a specific pyramid level $l$, DSM takes 1) query image feature maps $\mathbf{F}_{q,t}^l$ at time $t$ and $\mathbf{F}_{q,t-1}^l$ at time $t-1$; 2) scene image feature maps $\mathbf{F}_{s}^l$ with corresponding scene coordinates; 3) initial coordinate maps $\mathbf{\hat{D}}^{l}$. Then the DSM module predicts a coordinate map $\mathbf{D}^{l}$ by cost volume construction and coordinate regression. In the figure, $N$, $H$, $W$ is the number of scene images, image heights and image weights respectively. $K$ is the number of scene coordinates selected for regression and $d$ is the window size of candidate scene coordinates.
    }
    \label{fig:CRM}
\end{figure*}
\section{Related Work}

\noindent \textbf{Direct Pose Regression.} The prestigious PoseNet~\cite{kendall2015posenet} and its varients~\cite{kendall2015posenet,kendall2017geometric,brahmbhatt2018geometry,kendall2016modelling} regress the 6-DoF absolute poses directly from RGB images. These networks are trained in a supervised manner on RGB images with known ground truths by a regression loss of pose errors. 
Intuitively, these methods train a network to memorize the poses of all RGB images in a database.
It has been demonstrated in the work~\cite{sattler2019understanding} that direct pose regression yields results similar to pose approximation via image retrieval and the pose accuracy is usually inferior to the structure-based approaches, which are further classified into the following two categories.

\noindent \textbf{Sparse Feature Matching.} Methods~\cite{cheng2019cascaded, sattler2016efficient, sarlin2019coarse, taira2018inloc, middelberg2014scalable, middelberg2014scalable} based on sparse feature matching build 2D-3D correspondences by interest point detection~\cite{lowe2004distinctive, detone2018superpoint, dusmanu2019d2, bay2006surf, harris1988combined, mikolajczyk2004scale} and local descriptor matching~\cite{sarlin2019coarse, detone2018superpoint, sarlin2020superglue, dusmanu2019d2, lowe2004distinctive, calonder2010brief}. Then, poses are estimated by $PnP$ combined with RANSAC. To further improve the localization performance, the subsequent learning-based approaches gradually take a coarse-to-fine methodology~\cite{sarlin2019coarse,taira2018inloc,sarlin2018leveraging,mikolajczyk2004scale}. The other methods generally focus on improving the capability of local feature detectors~\cite{detone2018superpoint, dusmanu2019d2, savinov2017quad, zhang2018learning}, descriptors\cite{balntas2016learning, zhou2016evaluating, detone2018superpoint, dusmanu2019d2, zhang2018learning} and correspondence matching\cite{sarlin2020superglue}. A recent work along this direction is SuperGlue~\cite{sarlin2020superglue} and it achieves strong pose accuracy, especially in large-scale outdoor scenes. However, as limited by local feature descriptors, those methods tend to handle scenes with textureless regions or repeated patterns poorly. Instead, by leveraging global contexts, our method show better robustness on those scenes. Additionally, our method can generate dense coordinate maps which are important to various robotics and augmented reality applications.

\noindent \textbf{Dense coordinate regression.} Different from sparse 2D-3D correspondences, these methods directly regress the dense 3D scene coordinates of the query image and obtain the final camera pose by dense 2D-3D correspondences~\cite{brachmann2017dsac, schohn2000less, zhou2020kfnet, brachmann2019expert, shotton2013scene, li2020hierarchical}. Shotton et al.~\cite{shotton2013scene} proposes to regress the scene coordinates using a Random Forest. Along this direction, DSAC~\cite{brachmann2017dsac} and DSAC++~\cite{schohn2000less} employ convolutional neural networks to predict a dense coordinate map from a single RGB image.
KFNet~\cite{zhou2020kfnet} extends such ideology to the tasks of video sequence localization and embeds coordinate regression into Kalman filter within a deep learning framework.
It achieves the top performance on both single frame and video relocalization tasks.
Notably, all of those methods are scene-specific and cannot be generalized to arbitrary novel scenes, which limits their applications in scenarios requiring quick adaption to novel scenes.
SANet~\cite{yang2019sanet} is the first network proposed to regress coordinates in a scene-agnostic manner. However, it selects feature matches using ball query and uses Point-Net~\cite{qi2017pointnet, qi2017pointnet++} to regress 3D scene coordinates, which largely decreases coordinate accuracy and network efficiency. Our method is also scene agnostic, and we employ cost volumes to evaluate feature matches and compute 3D scene coordinates, which outperforms recent scene-specific and scene-agnostic methods including DSAC~\cite{brachmann2017dsac}, DSAC++~\cite{brachmann2019expert} and SANet~\cite{yang2019sanet}.

\noindent \textbf{Cost volume.} The proposed method in this paper is inspired by the ideology of cost volume which has been widely adopted in computer vision tasks, e.g. optical flow\cite{dosovitskiy2015flownet,sun2018pwc, ilg2017flownet, ranjan2017optical}, stereo matching~\cite{chang2018pyramid, mayer2016large, pang2017cascade} and multi-view stereo~\cite{yao2018mvsnet, han2015matchnet, vzbontar2016stereo}. Recent learning-based methods for optical flow or stereo matching extract feature pyramid, build cost volumes and make predictions in a coarse-to-fine manner~\cite{sun2018pwc, chang2018pyramid}. Since stereo matching or optical flow construct the cost volumes between image pairs, the number of costs for each pixel is fixed and can be arranged into a regular volume. On the other hand, multi-view stereo (MVS) build a dense 3D regular cost volume between images and the 3D space, with respect to a fixed number of depth or disparity hypothesis planes. However, the 3D dense cost volume used in MVS is infeasible to construct in our problem since it requires to sample a large number of hypothesis points, which makes the cost volume too large to process. Therefore, to build a regular 2D cost volume between a query image and a 3D scene efficiently, we propose a straightforward sorting strategy. The final dense coordinate maps are then obtained from the constructed cost volume. Thanks to the  cost volume based formulation, we can easily fuse temporal information to deal with video input. 

%% file: 3_1_overview.tex
\section{Method}
\subsection{Overview}
The overall framework of our system is illustrated in Fig.\ref{fig:pipeline}. The pipeline takes a single image or a video sequence as query input. For each query image, we first retrieve $N$ nearest scene images with corresponding coordinate maps via deep image retrieval~\cite{gordo2016deep}. Next, we extract a $L$-level feature pyramid for each query and scene image via the Feature Pyramid Network~\cite{lin2017feature}. In a coarse-to-fine manner, we then design a Dense Scene Matching (DSM) module at each pyramid level to regress the dense coordinate maps of gradually higher resolution and accuracy. Finally, the camera pose is estimated from the finest coordinate map by the standard RANSAC+PnP algorithm.

%% file: 3_2_pyramid.tex
\subsection{Feature and Coordinate Pyramid}\label{sec::feature_pyramid}
Given one query image $\mathbf{Q}_t$ at time $t$ and multiple reference scene images $\{\mathbf{S}_i| i = 1,...,n\}$, we generate a $L$-level pyramid of feature maps $\{\mathbf{F}^l| l = 1,...,L\}$ for each of them by ResNet50-FPN~\cite{lin2017feature}. We denote the query feature maps as $\mathbf{F}_{q}^l$ and the scene feature maps as $\mathbf{F}_{s}^l$. The feature vectors in $\mathbf{F}_{q}^l$ are referred as $\mathbf{f}_{q}^{l}$, and those in $\mathbf{F}_{s}^l$ are $\mathbf{f}_{s}^{l}$. The spatial size of feature maps at level $l$ is $H^l\times W^l$. 

For each scene image with known 3D coordinates, we also build a $L$-level coordinate pyramid $\{\mathbf{M}^l| l = 1,...,L\}$. The spatial size of each coordinate map is the same as that of the feature map $\mathbf{F}_{s}^l$. In order to deal with scenes at different scales, we transform the 3D scene coordinates to a local coordinate system, where the coordinates are normalized to zero-mean and unit standard deviation at all $x, y, z$ channels. 

We estimate the coordinate map in a coarse-to-fine manner. After initializing the coarsest level, the coordinate map $\mathbf{\hat{D}}^{l}$ at level $l$ is initialized by upsampling from $\mathbf{D}^{l+1}$.

%% file: 3_3_dense_matching.tex
\subsection{Dense Scene Matching}\label{sec:corr}
The overview of the DSM module is shown in Fig.~\ref{fig:CRM}.
At a specific level $l$, the input of DSM module includes: 1) query image feature maps $\mathbf{F}_{q}^l$; 2) scene image feature maps $\mathbf{F}_{s}^l$ and corresponding scene coordinate maps $M^l$; 3) initial coordinate maps $\mathbf{\hat{D}}^{l}$ upsampled from $\mathbf{D}^{l+1}$. The DSM module predicts the coordinate map $\mathbf{D}^{l}$ with more details from the initial $\mathbf{\hat{D}}^{l}$. Specifically, DSM consists of two steps, namely cost volume construction and coordinate regression. It first constructs a cost volume which measures the correlations between 2D query pixels and scene points (with known coordinates). It then regresses a dense coordinate map of the query image from the cost volume.

\subsubsection{Cost Volume Construction}\label{cost_vol}
This section explains the details of cost volume construction, which involves two processes, namely the scene correlation and temporal correlation.
The scene correlation measures similarity between query image pixels and scene points, while the temporal correlation measures the similarity between query image pixels from two neighboring frames in the query video clip.
Our network only uses scene correlation in training, and fuses both correlations at testing time.

\begin{figure}
    \centering
    \includegraphics[width=0.95\textwidth]{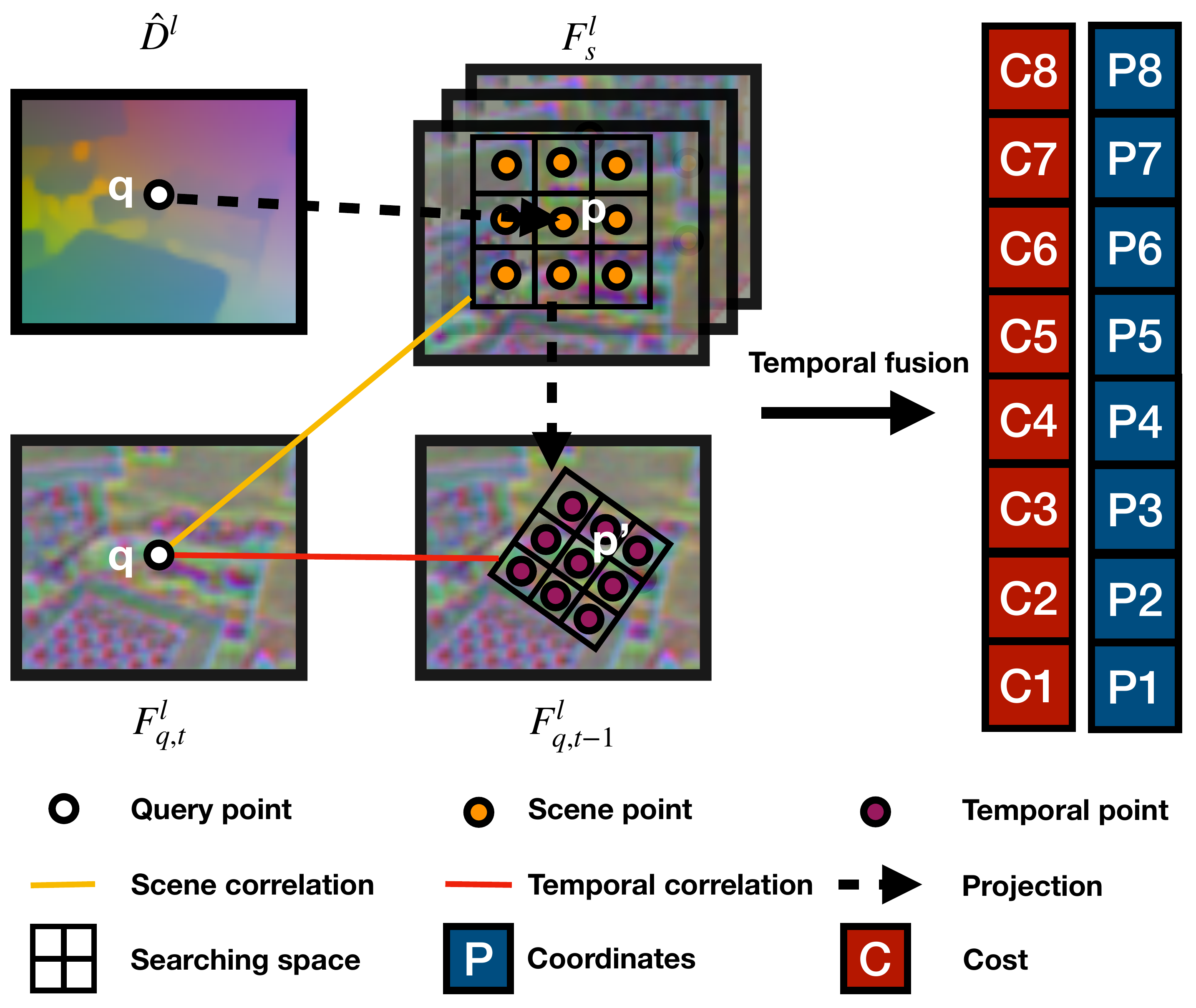}
    \caption{Demonstration of correlation fusion process. For a specific pixel $\mathbf{q}$ in $\mathbf{F}_{q,t}^l$, we obtain its scene correlation by projecting its corresponding 3D coordinate predicted in $\mathbf{\hat{D}}^l$ to a searching space of a $d\times d$ window in retrieved $N$ scene feature maps $\mathbf{F}_s^l$. Temporal correlation is obtained by projecting the scene coordinates within the searching space in $\mathbf{M}^l$ to $\mathbf{F}_{q,t-1}^l$. Finally, a $N \times d \times d$ correlation tensor is formed for each query pixel. 
    }
    \label{fig:Corr} 
\end{figure}

\noindent \textbf{Scene correlation.} 
The scene correlation is defined as cosine similarity between the features of query pixels and the ones of 3D scene points. We adopt a coarse-to-fine strategy in order to avoid the computation between all 2D-3D pairs. For the coarsest level, we compute the correlation between each query pixel and every 3D scene point since its initial depth is unknown. For the other levels, as shown in Fig.\ref{fig:Corr}, for an pixel $\mathbf{q}$ in the query feature map $\mathbf{F}_{q}^{l}$, we obtain its 3D coordinate from the initial coordinate map $\mathbf{\hat{D}}^{l}$. After that, we project the 3D coordinates to each scene image. Suppose the projected position is $\mathbf{p}$, we consider a $d\times d$ search window centered at $\mathbf{p}$ and compute the cosine similarity between the feature vector at $\mathbf{q}$ and those feature vectors for the pixels within the search window. In this way, we obtain a correlation vector of size $d\times d$ at the query pixel at $\mathbf{q}$. We initialize the correlation value as 0 if the corresponding position is out of the image. Given $N$ reference scene images, we obtain a $N\times d\times d$ scene correlations per pixel, which aggregate to a $H^l\times W^l \times (N\times d\times d)$ tensor, named correlation tensor. 

\noindent \textbf{Temporal correlation.} 
If the query input is a video sequence, we can leverage the result at the previous frame and the  correlation between neighboring video frames to enhance the result. Basically, if the camera pose is known, we can project a scene point $\mathbf{p}$ into the query video frame $\mathbf{Q}_{t-1}$ at $\mathbf{q}'$. Then the correlation between $\mathbf{p}$ and the query pixel $\mathbf{q}$ in video frame $\mathbf{Q}_{t}$ can be evaluated by the correlation between the two query pixels $\mathbf{q}'$ and $\mathbf{q}$. 

Specifically, we project all scene points to the query image $\mathbf{Q}_{t-1}$ according to the camera pose at $t-1$. Subsequently, we can compute the correlation between the feature vector at a query pixel in $\mathbf{Q}_{t}$ and the feature vectors of these projected pixels in $\mathbf{Q}_{t-1}$. In this way, for each query pixel, we also obtain a correlation vector of size $N\times d\times d$ by temporal correlation.

\noindent \textbf{Correlation fusion.} 
The final correlation score between a query pixel and a scene point is then computed from the scene correlation and temporal correlation by the equation, $Corr=\alpha Corr_s+(1-\alpha) Corr_t$, where $Corr_s$ stands for scene correlation and $Corr_t$ is temporal correlation. The parameter $\alpha$ balances $Corr_s$ and $Corr_t$. The hyper-parameter $\alpha$ is derived from the confidence score by $\alpha=min(s+0.4, 1)$, $s$ is the confidence score, which will be introduced in Sec.~\ref{Conf_esti}. Note that the fusion is applied to each of the $N$ reference scene images. At the end
of this fusion, we obtain a fused correlation tensor of size
$H^l\times W^l \times (N\times d\times d)$.

\noindent \textbf{Cost volume.}
We construct a cost volume by sorting the correlation values and selecting the top $K$ ($K=16$ in our implementation) scene coordinates. Although the sorting operation is not differentiable, the gradients can still be passed by the correlation values in the backward propagation during training. Intuitively, a higher correlation score means more accurate match between query pixels and scene points. 

After sorting, we obtain a cost volume of size $H^l \times W^l \times K$. We further concatenate this cost volume with the 3D scene coordinates of corresponding scene points to form a $H \times W \times 4K$ (1 for correlation and 3 for scene coordinates) cost-coordinate volume. This cost-coordinate volume is then processed by CNN to produce a dense coordinate map.

\subsubsection{Coordinate Regression}\label{CRM}
We design a network namely ${Net}_{coords}$ to estimate the final scene coordinate map by taking the input of cost volume, coordinate volume and image features. The cost-coordinate volume are first fed into a network consisting of $1\times 1$ convolutional layers  and produce a coordinate feature map. This coordinate feature map are concatenated with image feature map and fed into another network consisting of $3\times 3$ convolutional layers to predict the final coordiante map. The detailed architecture is illustrated in the supplementary material.

\subsubsection{Confidence Estimation}\label{Conf_esti}
In order to fuse the temporal correlations and scene correlations, we estimate a confidence value $s$ as the weighting parameter, as discussed in Sec. \ref{cost_vol}. We predict a certainty score for each pixel, which measures how accurate the coordinate prediction is. As illustrated in Fig.\ref{fig:CRM}, after predicting the coordinate map from only scene correlation, we concatenate it with the corresponding 2D pixel coordinates and feed to $Net_{conf}$ which outputs certainty scores. The architecture of $Net_{conf}$ is explained in supplementary material. We treat the certainty score estimation as a ranking problem. Coordinates with higher certainty scores are supposed to have smaller reprojection errors. This relation can be measured by the average precision metric. Therefore, we label each pixel as correct if its reprojection error of the estimated coordinate is smaller than a threshold (1 pixel in implementation) or incorrect otherwise and use average precision loss~\cite{revaud2019learning} to optimize $Net_{conf}$. The final confidence $s$ is the average certainty score over all pixels, and then the fusion score $\alpha$ can be computed. After fusing scene correlation and temporal correlation, we still use $Net_{coords}$ to predict the final coordinate map.

%% file: 3_4_loss.tex
\subsection{Training loss.} 
The total loss is the summation of the regression loss, $L_{regress}$, for coordinate regression and the average precision loss~\cite{revaud2019learning}, $L_{AP}$, for training certainty scores. For coordinate regression, we use $L_1$ distance errors between predicted coordinates and ground truth coordinates as training loss. 
\begin{align*}
    &L_{regress}=||Y_{coords}-\overline{Y}_{coords}||\\
    &L=L_{AP}+\frac{1}{n}\sum_{i=0}^n (L_{regress})\label{f:loss}
\end{align*}
Where $Y_{coords}$ is the absolute coordinate predicted from $Net_{coords}$, $\overline{Y}$ stands for the ground truth and $n$ is the number of query pixels.

%% file: table1.tex
\begin{table*}
\small
\centering
    \scalebox{0.95}{\begin{tabular}{c|c|c|c|c|c|c|c|c}
    \hline
         &7scenes (Indoor)&Chess&Fire&Heads&Office&Pumpkin&Kitchen&Stairs\\
         
         \hline
         {\multirow{3}{*}{\rotatebox[origin=c]{90}{Sparse}}}&Active Search&1.96$^{\circ}$, 0.04m&1.53$^{\circ}$, 0.03m&1.45$^{\circ}$, 0.02m& 3.61$^{\circ}$, 0.09m&3.10$^{\circ}$, 0.08m&3.37$^{\circ}$, 0.07m&2.22$^{\circ}$, \textbf{0.03m}\\
         
         &InLoc&1.05$^{\circ}$, 0.03m&1.07$^{\circ}$, 0.03m&0.16$^{\circ}$, 0.02m&1.05$^{\circ}$, 0.03m&1.55$^{\circ}$, 0.05m&1.31$^{\circ}$, 0.04m&2.47$^{\circ}$, 0.09m\\
         
         &HLoc&\textbf{0.79$^{\circ}$}, \textbf{0.02m}&\textbf{0.87$^{\circ}$}, \textbf{0.02m}&\textbf{0.92$^{\circ}$}, \textbf{0.02m}&\textbf{0.91$^{\circ}$}, \textbf{0.03m}&\textbf{1.12$^{\circ}$}, \textbf{0.05m}&\textbf{1.25$^{\circ}$}, \textbf{0.04m}&\textbf{1.62$^{\circ}$}, 0.06m\\
         \hline
         {\multirow{6}{*}{\rotatebox[origin=c]{90}{Dense}}}&DSAC(*)&0.7$^{\circ}$, 0.02m&1.0$^{\circ}$, 0.03m&1.3$^{\circ}$, 0.02m&1.0$^{\circ}$, 0.03m&1.3$^{\circ}$, 0.05m&1.5$^{\circ}$, 0.0.5m&49.4$^{\circ}$, 1.9m\\
         
         &DSAC++(*)&\textbf{0.5$^{\circ}$}, \textbf{0.02m}&0.9$^{\circ}$, \textbf{0.02m}&\textbf{0.8$^{\circ}$}, \textbf{0.01m}&0.7$^{\circ}$, \textbf{0.03m}&1.1$^{\circ}$, \textbf{0.04m}&1.1$^{\circ}$, 0.04m&2.6$^{\circ}$, 0.09m\\
         
         &KFNet(*)&0.65$^{\circ}$, \textbf{0.02m}&0.9$^{\circ}$, \textbf{0.02m}&0.82$^{\circ}$, \textbf{0.01m}&\textbf{0.69$^{\circ}$}, \textbf{0.03m}&\textbf{1.02$^{\circ}$}, \textbf{0.04m}&1.16$^{\circ}$, 0.04m&\textbf{0.94$^{\circ}$}, \textbf{0.03m}\\
         
         &SANet&0.88$^{\circ}$, 0.03m&1.10$^{\circ}$, 0.03m&1.48$^{\circ}$, 0.02m&1.03$^{\circ}$, 0.03m&1.32$^{\circ}$, 0.05m&1.4$^{\circ}$, 0.04m&4.59$^{\circ}$, 0.16m\\
      
         &Ours (Single)&0.71$^{\circ}$, 0.02m&0.85$^{\circ}$, 0.02m&0.85$^{\circ}$, 0.01m&0.84$^{\circ}$, 0.03m&1.16$^{\circ}$, 0.04m&1.17$^{\circ}$,0.04m&1.33$^{\circ}$, 0.05m\\
         
         &Ours (Video)&0.68$^{\circ}$, \textbf{0.02m}&
         \textbf{0.80}$^{\circ}$, \textbf{0.02m}&\textbf{0.80}$^{\circ}$, \textbf{0.01m}&0.78$^{\circ}$, \textbf{0.03m}&1.11$^{\circ}$, \textbf{0.04m}&\textbf{1.11$^{\circ}$},\textbf{0.03m}&1.16$^{\circ}$, 0.04m\\
         \hline
         \hline
        
        \multicolumn{9}{c}{\begin{tabular}{c|c|c|c|c|c|c|c}
             &Cambridge (outdoor)&Great Court&King’s College&Old Hospital&Shop Facade&St. Mary’s Church&Street\\
             \hline
             \parbox{2mm}{\multirow{3}{*}{\rotatebox[origin=c]{90}{ Sparse}}}&Active Search&0.6$^{\circ}$, 1.20m&0.6$^{\circ}$, 0.42m&1.0$^{\circ}$, 0.44m&0.4$^{\circ}$, 0.12m&0.5$^{\circ}$, 0.19m&\textbf{0.8$^{\circ}$}, 0.85m\\
             &InLoc&0.62$^{\circ}$, 1.20m&0.82$^{\circ}$, 0.46m&0.96$^{\circ}$, 0.48m&0.50$^{\circ}$, 0.11m&0.63$^{\circ}$, 0.18m&2.16$^{\circ}$, 0.75m\\
             &HLoc&\textbf{0.21$^{\circ}$}, \textbf{0.38m}&\textbf{0.31$^{\circ}$}, \textbf{0.17m}&\textbf{0.39$^{\circ}$}, \textbf{0.23m}&\textbf{0.37$^{\circ}$}, \textbf{0.07m}&\textbf{0.29$^{\circ}$}, \textbf{0.10m}&1.32$^{\circ}$, \textbf{0.62m}\\
            \hline
             \parbox{2mm}{\multirow{6}{*}{\rotatebox[origin=c]{90}{ Dense}}}&DSAC(*)&1.5$^{\circ}$, 2.8m&0.5$^{\circ}$, 0.30m&0.6$^{\circ}$, 0.33m&0.4$^{\circ}$, 0.09m&1.6$^{\circ}$, 0.55m&\\

             &DSAC++(*)&0.2$^{\circ}$, \textbf{0.40m}&0.3$^{\circ}$, 0.18m&0.3$^{\circ}$,0.2m&\textbf{0.3$^{\circ}$}, 0.06m&0.4$^{\circ}$, 0.13m&\\
            &KFNet(*)&0.21$^{\circ}$, 0.42m&\textbf{0.27$^{\circ}$}, \textbf{0.16m}&\textbf{0.28$^{\circ}$}, \textbf{0.18m}&0.35$^{\circ}$, \textbf{0.05m}&0.35$^{\circ}$, 0.12m&\\
            &SANet&1.95$^{\circ}$, 3.28m&0.42$^{\circ}$, 0.32m&0.53$^{\circ}$, 0.32m&0.47$^{\circ}$, 0.10m&0.57$^{\circ}$, 0.16m&12.64$^{\circ}$, 8.74m\\
            &Ours (Single)&0.23$^{\circ}$, 0.44m&0.36$^{\circ}$, 0.19m&0.39$^{\circ}$, 0.24m&0.38$^{\circ}$, 0.07m&0.35$^{\circ}$, 0.12m&1.71$^{\circ}$, 0.68m\\
             &Ours (Video)& \textbf{0.19$^{\circ}$}, 0.43m&0.35$^{\circ}$, 0.19m&0.38$^{\circ}$, 0.23m&\textbf{0.30$^{\circ}$}, 0.06m&\textbf{0.34$^{\circ}$}, \textbf{0.11m}&\textbf{1.53$^{\circ}$}, \textbf{0.61m}\\
             \hline


           
        \end{tabular}}\\
        \hline
    \end{tabular}}
    \caption{Performance comparison in terms of rotation errors ($^{\circ}$) and translation errors (m). (*) indicates scene-specific methods.}
    \label{tab:local}
\end{table*}

%% file: table2.tex
\begin{table*}
\small
    \centering
    \scalebox{0.9}{\begin{tabular}{c||c|c|c|c|c|c|c}
        \hline
        
         &&\multicolumn{3}{|c}{Single frame localization}&\multicolumn{3}{|c}{Video localization}  \\
        \hline
         &Acc. thresh& Median&Mean&Acc.& Median&Mean&Acc.\\
         \hline
        \hline
         Chess&5$^{\circ}$, 0.05&0.713$^{\circ}$, 0.021& 0.824$^{\circ}$, 0.024&94.5 &0.684$^{\circ}$, 0.020&0.795$^{\circ}$, 0.023&96.1 (+1.6)\\
        \hline
        Fire&5$^{\circ}$, 0.05& 0.856$^{\circ}$, 0.021&1.025$^{\circ}$, 0.027&93.8& 0.802$^{\circ}$, 0.020&0.878$^{\circ}$, 0.020&94.5 (+0.7)\\
        \hline
        Heads&5$^{\circ}$, 0.05& 0.846$^{\circ}$, 0.013&1.369$^{\circ}$, 0.023&96.4& 0.802$^{\circ}$, 0.013&0.957$^{\circ}$, 0.016&99.5 (+3.1)\\
        \hline
        Office&5$^{\circ}$, 0.05&0.843$^{\circ}$, 0.028&0.983$^{\circ}$, 0.037&82.3&0.782$^{\circ}$, 0.026&0.937$^{\circ}$, 0.034&84.2 (+1.9)\\
        \hline
        Pumpkin&5$^{\circ}$, 0.05&1.164$^{\circ}$, 0.043&2.224$^{\circ}$, 0.112&57.0&1.113$^{\circ}$, 0.043&1.823$^{\circ}$, 0.083&57.2 +(0.2)\\
        \hline
        Kitchen&5$^{\circ}$, 0.05&1.165$^{\circ}$, 0.038&3.145$^{\circ}$, 0.082&68.7&1.115$^{\circ}$, 0.034&1.358$^{\circ}$, 0.044&69.2 (+0.5)\\
        \hline
        Stairs&5$^{\circ}$, 0.05&1.356$^{\circ}$, 0.045&3.424$^{\circ}$, 0.197&53.9&1.157$^{\circ}$, 0.037&1.553$^{\circ}$, 0.069&69.9 (+16.0)\\
        \hline
        \hline
        Great Court&5$^{\circ}, 1.0$&0.209$^{\circ}$, 0.444&6.043$^{\circ}$, 5.624&68.5&
        0.193$^{\circ}$,0.428&4.023$^{\circ}$, 4.017&76.7 (+8.2)\\
        \hline
        King’s College&5$^{\circ}, 0.5$&0.358$^{\circ}$, 0.194&0.574$^{\circ}$, 0.424&82.9&
        0.353$^{\circ}$, 0.188&0.522$^{\circ}$, 0.367&84.6 (+1.7)\\
        \hline
        Old Hospital&5$^{\circ}, 0.3$&0.388$^{\circ}$, 0.243&0.387$^{\circ}$, 0.502&41.2&
        0.382$^{\circ}$, 0.228&0.372$^{\circ}$, 0.498&43.7 (+2.5)\\
        \hline
        Shop Facade&5$^{\circ}, 0.2$&0.375$^{\circ}$, 0.074&0.623$^{\circ}$, 0.131&84.2&0.303$^{\circ}$, 0.061&0.574$^{\circ}$, 0.112&86.4 (+2.2)\\
        \hline
        St. Mary’s Church&5$^{\circ}, 0.3$&0.353$^{\circ}$, 0.118&1.146$^{\circ}$, 0.374&91.4&0.342$^{\circ}$, 0.111&0.845$^{\circ}$, 0.264&93.7 (+2.3)\\
        \hline
        Street&5$^{\circ}, 2.0$&1.711$^{\circ}$, 0.684&22.551$^{\circ}$, 27.111&62.2&
        1.523$^{\circ}$, 0.609&20.756$^{\circ}$, 25.862&64.8 (+2.6)\\
        \hline
    \end{tabular}}
    \caption{Comparison of single frame based localization and video-based localization. For 7scenes, we use common threshold (5$^{\circ}$, 0.05m) to calculate accuracy. We can see video-based has significantly lower mean errors and higher accuracy.
    }
    \label{tab:video}
\end{table*}

%% file: experiment.tex
\section{Experiments}
\subsection{Experiment Settings}\label{experiment setting}
\noindent \textbf{Dataset.} We evaluate our method on both the indoor dataset 7scenes~\cite{glocker2013real} and the outdoor dataset Cambridge Landmarks~\cite{kendall2015posenet}. For 7scenes, it contains 7 different scenes with raw RGB-D video sequences captured by a handheld Kinect RGB-D camera. It also provides camera poses and a dense 3D model for each scene generated by KinectFusion~\cite{izadi2011kinectfusion}. Cambridge Landmarks dataset contains 6 different outdoor scenes with RGB video frames labelled with full 6-DOF camera poses. We train our network using ScanNet dataset~\cite{dai2017scannet}, which is a RGB-D video dataset consisting of 2.5M views in 1513 scenes annotated with 3D camera poses and dense depth maps.

\noindent \textbf{Data processing.}
All the images of the 7scenes~\cite{glocker2013real}, Cambridge Landmarks~\cite{kendall2015posenet} and ScanNet~\cite{dai2017scannet} datasets are downsized to $384\times 512$. To form the training data, we first randomly sample about 160k images from ScanNet dataset as query images. For each query image, we retrieve $5$ and $10$ corresponding scene images in the same video sequence for training and testing respectively by the learning-based image retrieval approach~\cite{gordo2016deep}. In order to encourage query-scene image pairs with different viewing angles, we only keep the scene images of the same video sequence that are at least 50 frames away from a given query image. We follow the multi-view stereo reconstruction method adopted in the DSAC~\cite{brachmann2017dsac} to obtain dense 3D coordinates of the Cambridge Landmarks.

\noindent \textbf{Training.}
We only use Scannet as training data for the inference on 7scenes dataset. As for a specific scene of the outdoor dataset Cambridge Landmarks, we fine-tune our pretrained model with the other 5 scenes. ResNet50-FPN~\cite{lin2017feature} is regarded as our backbone network for all the following experiments. Our model is trained with an AdamW optimizer~\cite{loshchilov2017decoupled}, whose base learning rate is 0.0005, and a batch size of 16 in a single RTX TITAN GPU for 50000 iterations.

\begin{figure*}
    \centering 
  \includegraphics[width=0.95\textwidth]{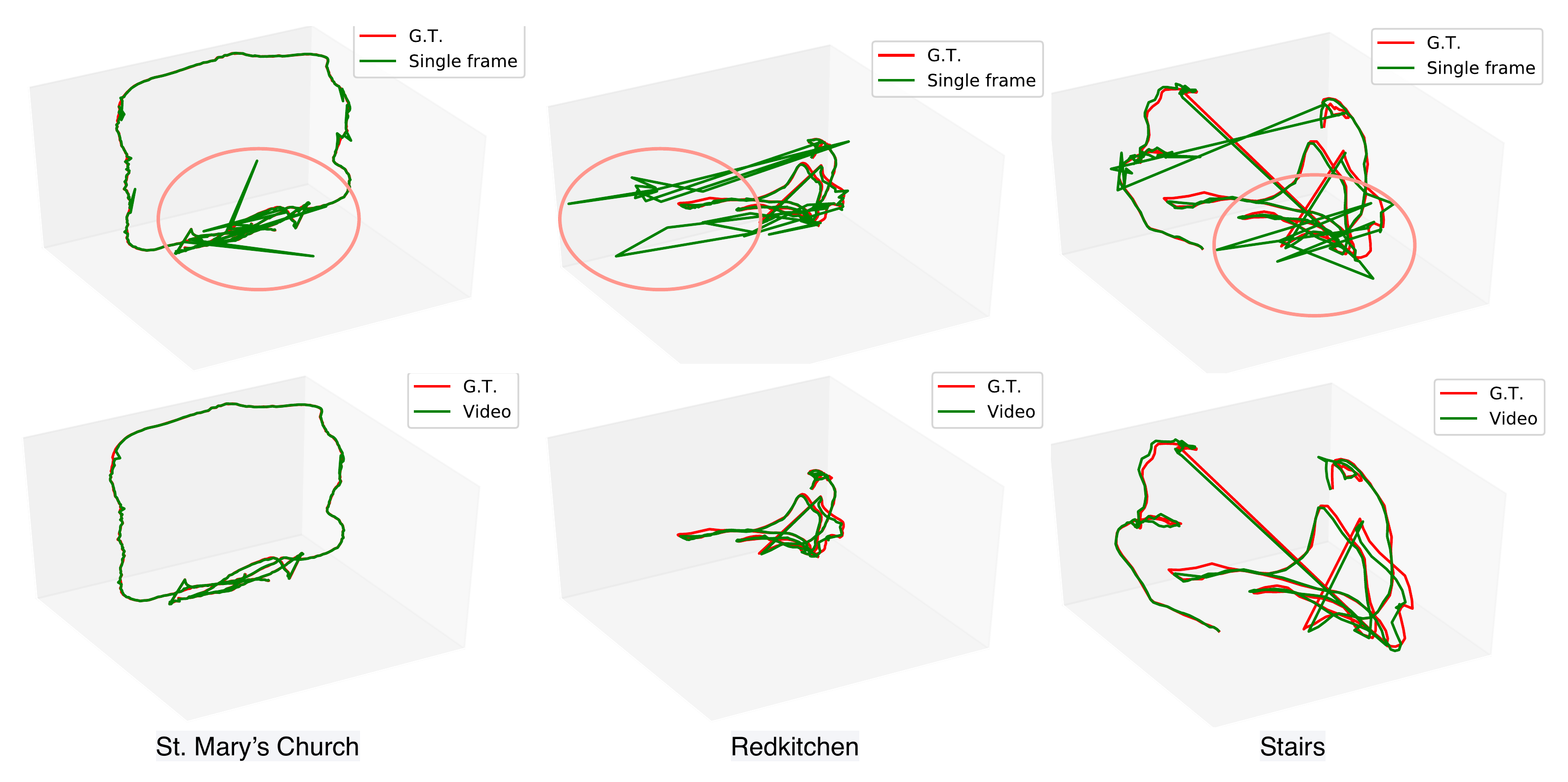}\hfill
\caption{The comparison of camera trajectories between the single frame (first row) and video localization (second row) via the proposed dense scene matching network. The visualized results are respectively \textit{Redkitchen} and \textit{Stairs} in 7-scenes dataset, and \textit{St. Mary’s Church} sequence in Cambridge Landmarks dataset. In the first row, the outliers are shown in the red circles.}\label{fig:trajectory}
\end{figure*}



\subsection{Localization Accuracy}\label{localize_acc}
In this section, we mainly compare our approach with two classes of methods, namely sparse feature matching~\cite{sattler2012improving,taira2018inloc,sarlin2019coarse} and dense coordinate regression methods~\cite{brachmann2017dsac,schohn2000less,zhou2020kfnet,yang2019sanet}. We measure localization accuracy in terms of median errors in translation and rotation. As shown in Table.~\ref{tab:local}, the proposed DSM approach achieves state-of-the-art performance among both sparse matching and dense regression methods. 

Compared with sparse matching methods, the pose accuracy of our approach is superior to that of Active Search~\cite{sattler2012improving} and InLoc~\cite{taira2018inloc}. HLoc~\cite{sarlin2019coarse}, upgraded with SuperPoint~\cite{detone2018superpoint} for feature detection and Superglue~\cite{sarlin2020superglue} for feature correspondence matching, is considered and such upgrade brings higher relocalization accuracy compared with the original HLoc approach~\cite{sarlin2019coarse} as reported in the work~\cite{sarlin2020superglue}. We can see that DSM outperforms HLoc in 7scenes, and it is slightly inferior to HLoc in outdoor Cambridge Landmarks dataset which contains much more salient texture for sparse feature matching. 

When comparing with scene-specific dense coordinate regression methods, the proposed scene-agnostic approach DSM outperforms DSAC~\cite{brachmann2017dsac} by a large margin and obtains slightly superior performance than DSAC++~\cite{schohn2000less}. Even for KFNet~\cite{zhou2020kfnet} with the top performance on single frame and video localization tasks, our approach achieves comparable performance. In comparison with the scene-agnostic SANet~\cite{yang2019sanet}, DSM shows obvious superior performance. 

Table.\ref{tab:video} shows the detailed comparison of metrics of median errors, mean errors and the pose accuracy falling within certain accuracy threshold (\textit{Acc. thresh}) between single frame localization and video localization methods. As shown, after applying the temporal fusion, the localization accuracy notably increases indeed. In addition, Fig. \ref{fig:trajectory} shows the trajectories of \textit{Redkitchen} and \textit{Stair} sequence of 7-scenes dataset and \textit{St. Mary’s Church} sequence of Cambridge dataset. We can see that the trajectories of our single frame localization contains some outliers while our video localization is able to remove most of them.

\begin{figure}
    \includegraphics[width=\textwidth]{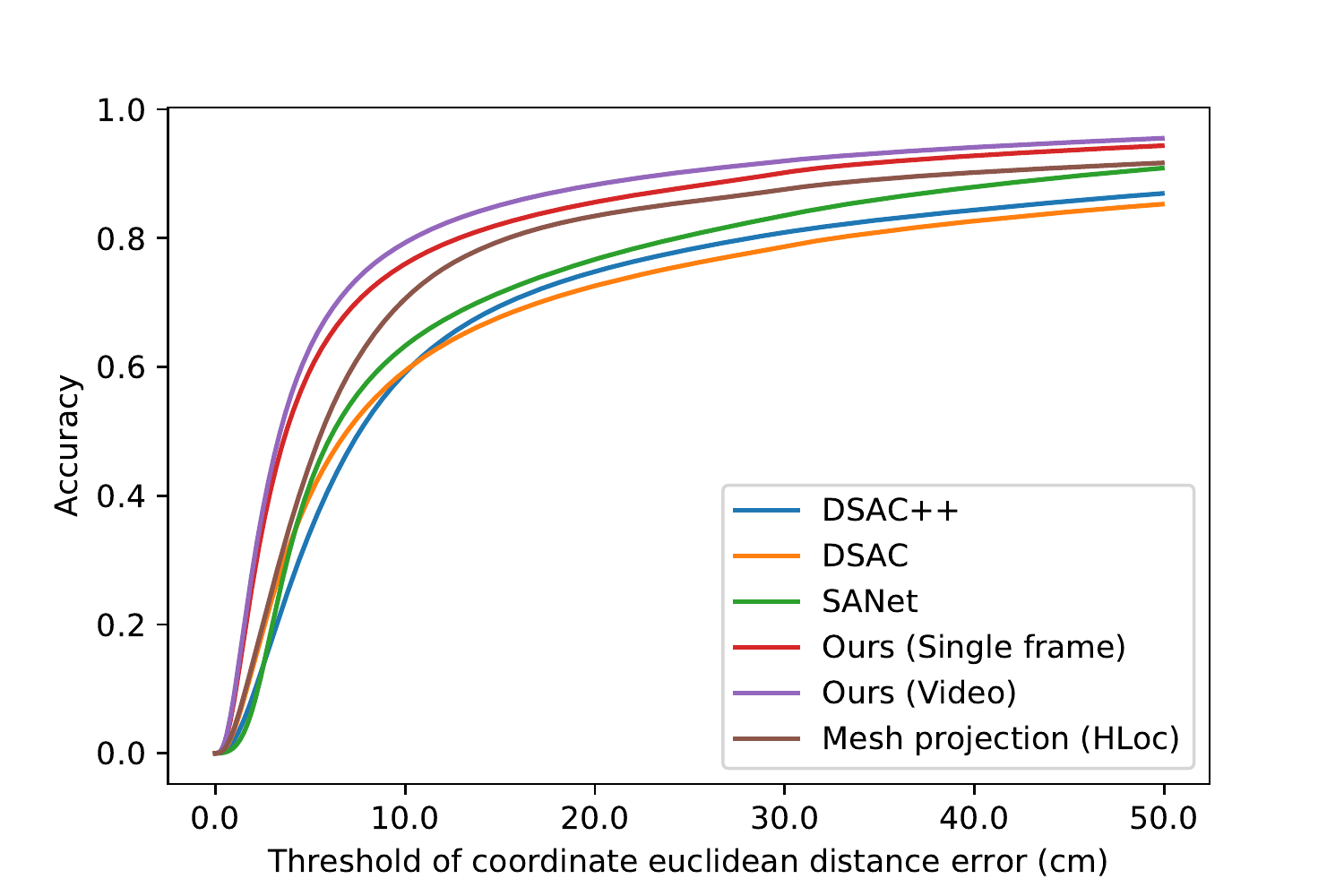}
    \caption{The comparison of cumulative distribution functions of scene coordinate errors between different localization approaches.}
    \label{fig:coords_acc}
\end{figure}
\begin{figure*}
    \centering 
\begin{subfigure}{0.18\textwidth}
  \includegraphics[width=\linewidth]{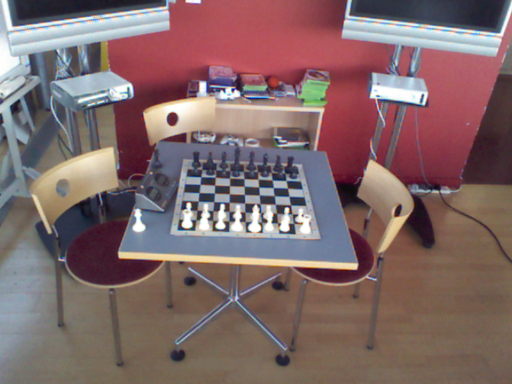}
\end{subfigure}\hfil 
\begin{subfigure}{0.18\textwidth}
  \includegraphics[width=\linewidth]{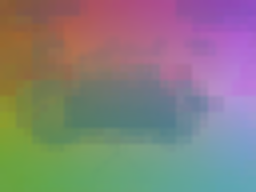}
\end{subfigure}\hfil 
\begin{subfigure}{0.18\textwidth}
  \includegraphics[width=\linewidth]{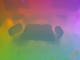}
\end{subfigure}\hfil 
\begin{subfigure}{0.18\textwidth}
  \includegraphics[width=\linewidth]{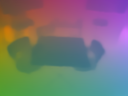}
\end{subfigure}\hfil 
\begin{subfigure}{0.18\textwidth}
  \includegraphics[width=\linewidth]{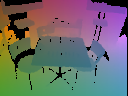}
\end{subfigure}\hfil 

\medskip
\begin{subfigure}{0.18\textwidth}
  \includegraphics[width=\linewidth]{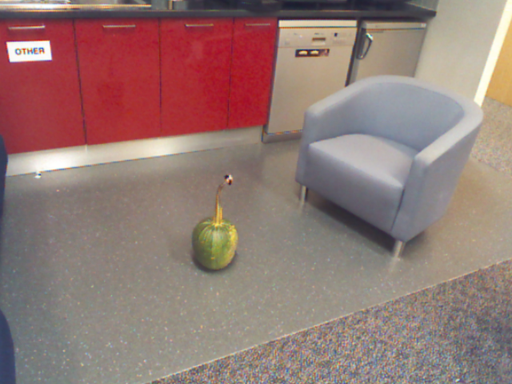}
  \caption{Query}
\end{subfigure}\hfil 
\begin{subfigure}{0.18\textwidth}
  \includegraphics[width=\linewidth]{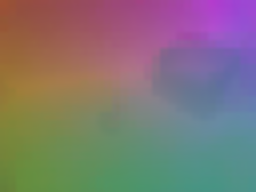}
  \caption{SANet}\label{fig:2}
\end{subfigure}\hfil 
\begin{subfigure}{0.18\textwidth}
  \includegraphics[width=\linewidth]{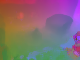}
   \caption{DSAC++}
\end{subfigure}\hfil 
\begin{subfigure}{0.18\textwidth}
  \includegraphics[width=\linewidth]{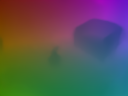}
  \caption{DSM}\label{DSM_A}
\end{subfigure}\hfil 
\begin{subfigure}{0.18\textwidth}
  \includegraphics[width=\linewidth]{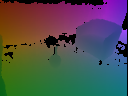}
  \caption{G.T.}
\end{subfigure}\hfil 

\caption{Coordinate map visualization for SANet, DSAC++ and DSM.}
\label{fig:coords_vis}
\end{figure*}

\begin{table}[]
    \centering
    \begin{tabular}{c|c|c}
    \hline
         &Run time&GPU memory usage  \\
        \hline
         SANet&0.33s &5GB \\
         \hline
         Ours&0.21s &2.7GB\\
         \hline
    \end{tabular}
    \caption{Efficiency comparison of SANet and DSM.}
    \label{tab:efficiency}
\end{table}

\subsection{Scene coordinate accuracy}
In terms of scene coordinate accuracy, we compare our method with SANet~\cite{yang2019sanet}, DSAC~\cite{brachmann2017dsac}, DSAC++~\cite{schohn2000less} and HLoc~\cite{sarlin2019coarse} on the whole 7scenes dataset. Since HLoc cannot directly output a dense coordinate map, we first get dense depth maps by projecting reconstructed mesh to its predicted poses and compute coordinates by back-projection. We calculate the coordinate accuracy under different euclidean distance error threshold and plot cumulative distribution function in Fig.\ref{fig:coords_acc}. We can see that the accuracy of coordinate maps from our network outperforms SANet, DSAC and DSAC++ by a large margin. More specifically, we surpass SANet by 16\% and DSAC++ by 20\% when the threshold is set to 10 cm. The projected coordinates of HLoc is more accurate than DSAC, DSAC++ and SANet, but is under-performed by DSM. In addition, our temporal-based coordinate map regression boosts accuracy compared with our single frame prediction.

We also visualize coordinate map in Fig.\ref{fig:coords_vis} for DSM, SANet and DSAC++. In general, the coordinate map produced by DSM has higher quality and preserves more details than SANet and DSAC++. SANet randomly sample coordinates from search space with ball query, and the best match may be dropped due to this operation. As a result, its coordinate maps contain a large number of artifacts. DSAC++ is able to produce coordinate map with more details, but artifacts exist in some regions as well.

\subsection{Efficiency}
Table.\ref{tab:efficiency} shows the running time and GPU memory usage to localize a single query frame with 5 scene images.
We list the statistics of SANet since it is the only localization pipeline that predict dense coordinate map for an arbitrary novel scenes.
Here, the image retrieval time is not included.
Compared with SANet, Our network reduces the time consumption by $33\%$ and memory consumption by $46\%$. The efficiency can be further improved by adapting light-weight backbones.

\begin{figure}
    \centering
    \begin{subfigure}{\textwidth}
      \includegraphics[width=\textwidth]{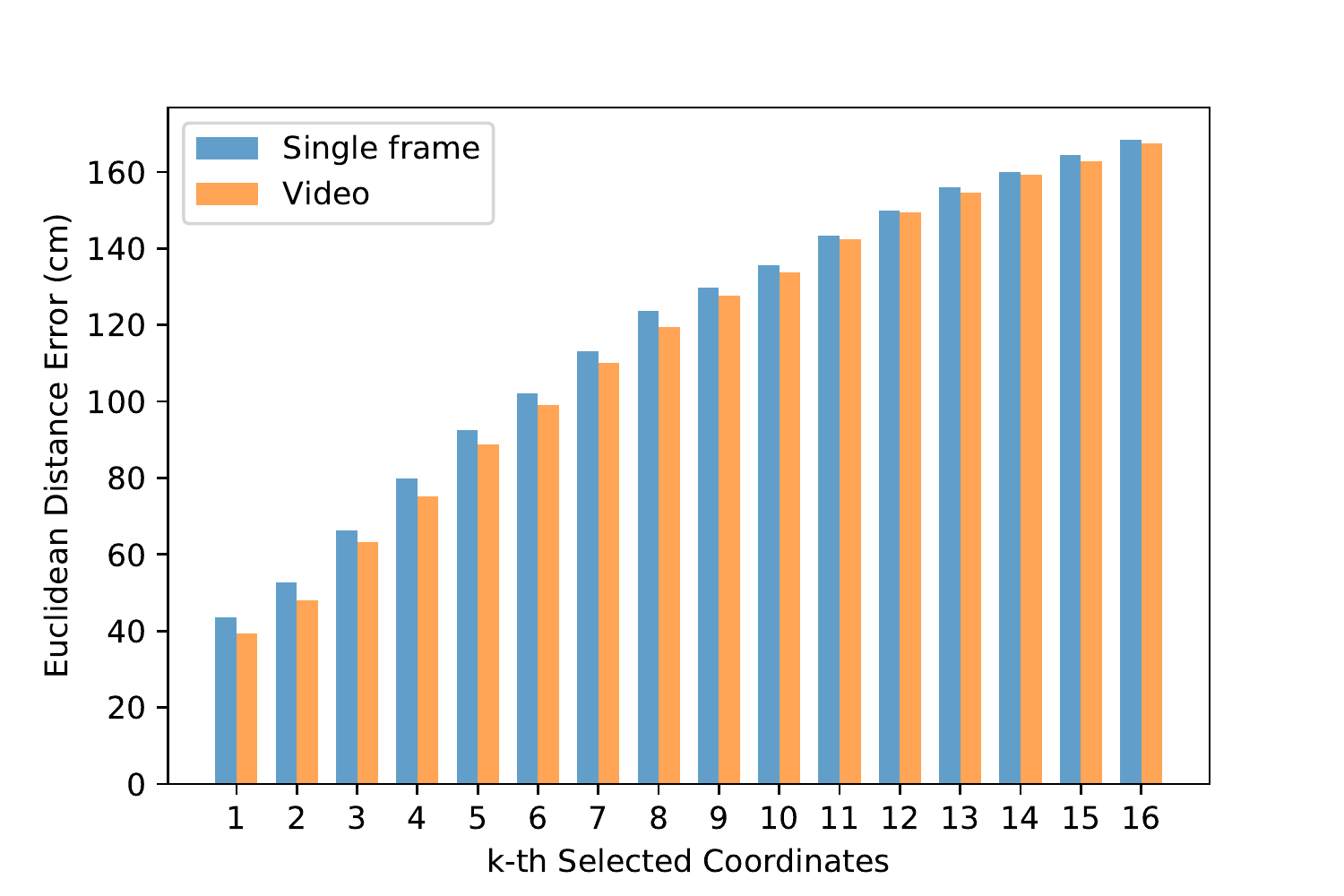}
      
    \end{subfigure}
    
    \caption{The average scene coordinate errors of the k-th selected coordinates with respect to ground truth.}
    \label{fig:euc_err}
\end{figure}

\subsection{Analysis of correlation}
Our proposed approach assumes that high query-scene correlations lead to more accurate corresponding scene coordinates for query pixels. To verify this argument, we evaluate the relationship between the correlation and scene coordinate errors with respect to ground truth. For each pixel in the query image, we select the top K scene coordinate candidates in $5^{th}$ level of coordinate pyramid.  Then for each ranking index $k$, we take the average of the euclidean distance error between selected scene coordinates and ground truths over all query pixels. Finally, We define the $k^{th}$ average scene coordinate error as $e_k=\frac{1}{n}\sum_{i=1}^{n} \sqrt{||Y_{k}^{i}-\overline{Y}||_2}$, where $n$ is the number of pixels in a query image, for the $i^{th}$ query pixel, $Y_{k}^{i}$ is the $k^{th}$ corresponding scene coordinate and $\overline{Y}$ is the ground truth. We summarize the statistics in 7scenes dataset and plot $e_k$ in Fig.\ref{fig:euc_err}. It can be seen that the scene coordinate error gradually becomes larger when correlation becomes smaller. In other words, high correlation stands for more accurate scene coordinate selection for a specific query pixel. In addition, we also include the evaluation for temporal-based model, which obtains consistently lower euclidean distance errors than single frame model, indicating that correlation fusion further improves the accuracy of selected scene coordinates.

%% file: conclusion.tex
\section{Conclusion}
In this paper, we present dense scene matching (DSM) for visual localization. DSM is able to estimate dense coordinate maps for arbitrary novel scenes.
First, DSM builds a cost volume between a query image and a scene by sorting and selecting the top K highest correlations per pixel.
Then, the cost volume with the corresponding coordinates are feed into CNN for dense coordinate regression and a temporal fusion module is introduced to further improve the accuracy of the dense coordinate map.
Finally, the camera poses are then estimated by $PnP$ together with RANSAC algorithms.
We demonstrated the effectiveness of DSM on both indoor and outdoor datasets. This scene-agnostic method yields comparable accuracy among all scene-specific methods and outperforms scene-agnostic methods in terms of both localization and coordinate accuracy. 